\documentclass{article}

\usepackage{times}
\usepackage{graphicx} 

\usepackage{natbib}

\usepackage{algorithm}
\usepackage{algorithmic}

\usepackage{hyperref}

\usepackage[accepted]{icml2017}
\usepackage{amsmath}
\usepackage{amsfonts}
\usepackage{subcaption}
\usepackage{booktabs}
\usepackage{comment}
\usepackage{multirow}
\usepackage{color}
\usepackage{bbm}

\icmltitlerunning{Image-to-Markup Generation with Coarse-to-Fine Attention}

\begin{document} 

\twocolumn[
\icmltitle{Image-to-Markup Generation with Coarse-to-Fine Attention}



\icmlsetsymbol{equal}{*}

\begin{icmlauthorlist}
\icmlauthor{Yuntian Deng}{harvard}
\icmlauthor{Anssi Kanervisto}{finland}
\icmlauthor{Jeffrey Ling}{harvard}
\icmlauthor{Alexander M. Rush}{harvard}
\end{icmlauthorlist}

\icmlaffiliation{harvard}{Harvard University}
\icmlaffiliation{finland}{University of Eastern Finland}

\icmlcorrespondingauthor{Yuntian Deng}{dengyuntian@seas.harvard.edu}

\icmlkeywords{deep learning, attention model, image captioning}

\vskip 0.3in
]



\printAffiliationsAndNotice{}  

\begin{abstract} 
  We present a neural encoder-decoder model to convert images into
  presentational markup based on a scalable coarse-to-fine attention
  mechanism. Our method is evaluated in the context of image-to-LaTeX
  generation, and we introduce a new dataset of real-world rendered
  mathematical expressions paired with LaTeX markup.  We show that
  unlike neural OCR techniques using CTC-based models, attention-based
  approaches can tackle this non-standard OCR task. Our approach
  outperforms classical mathematical OCR systems by a large margin on
  in-domain rendered data, and, with pretraining, also performs well
  on out-of-domain handwritten data. To reduce the inference
  complexity associated with the attention-based approaches, we
  introduce a new coarse-to-fine attention layer that selects a
  support region before applying attention. 
\end{abstract}

\begin{figure*}
  \centering
  \includegraphics[width=\linewidth]{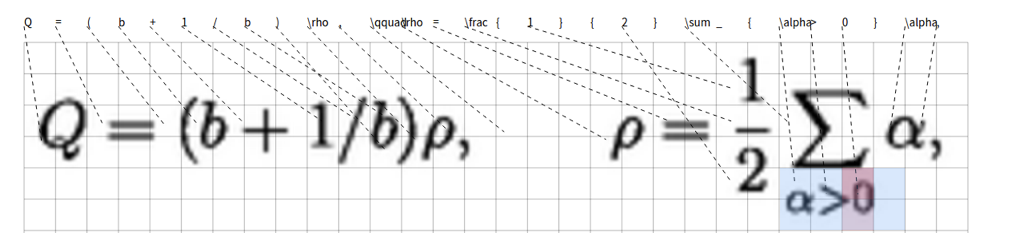}
  \vspace*{-0.5cm}
  \caption{ \small Example of the model generating mathematical
    markup. The model generates one LaTeX symbol $y$ at a time based
    on the input image $\mathbf{x}$. The gray lines highlight the
    $H\times V$ grid features $\mathbf{{V}}$ formed by the row encoder from the CNN's output. The dotted lines indicate the center
    of mass of $\mathbf{\alpha}$ for each token (only non-structural tokens are
    shown). The blue cell indicates the support set selected by the coarse-level attention for the symbol ``0'', while the red cells indicate its fine-level attention. White space around the image has been trimmed for visualization. The actual size of the blue mask is $4\times 4$. See \url{http://lstm.seas.harvard.edu/latex/} for a complete interactive
    version of this visualization over the test set. }
  \label{fig:visattn}
  \vspace*{-0.2cm}
\end{figure*}

\section{Introduction}

Optical character recognition (OCR) is most commonly used to recognize
natural language from an image; however, as early as the work of
\citet{anderson1967syntax}, there has been research interest in
converting images into structured language or \textit{markup} that
defines both the text itself and its presentational semantics. The
primary target for this research is OCR for mathematical expressions, and
how to handle presentational aspects such as sub and superscript
notation, special symbols, and nested fractions
\cite{belaid1984syntactic,DBLP:journals/ijdar/ChanY00}. The most
effective systems combine specialized character segmentation with
grammars of the underlying mathematical layout language \cite{miller1998ambiguity}. A prime
example of this approach is the \textsc{Infty} system that is used to
convert printed mathematical expressions to LaTeX and other markup
formats \cite{suzuki2003infty}. Other, mostly proprietary systems,
have competed on this task as part of the CROHME handwritten mathematics 
challenge~\cite{mouchere2013icdar,mouchere2014icfhr}.

Problems like OCR that require joint processing of image and text data
have recently seen increased research interest due to the refinement
of deep neural models in these two domains. For instance, advances
have been made in the areas of handwriting recognition
\cite{ciresan2010deep}, OCR in natural scenes
\cite{jaderberg2015deep,jaderberg2016reading,wang2012end} and image
caption generation~\cite{karpathy2015deep,vinyals2015show}. At a
high-level, each of these systems learn an abstract encoded
representation of the input image which is then decoded to generate a
textual output. In addition to performing quite well on standard
tasks, these models are entirely data driven, which makes them
adaptable to a wide range of datasets without requiring heavy preprocessing or domain specific engineering.

However, we note that tasks such as image captioning differ from the
traditional mathematical OCR task in two respects: first, unlike image captioning, the traditional OCR task assumes a left-to-right ordering, so neural systems addressing this problem have primarily relied on
Connectionist Temporal Classification (CTC)
\cite{graves2006connectionist} or stroke-based approaches. Second, the image captioning task theoretically allows for systems to focus
their attention anywhere, and thus does not directly test a system's ability to maintain consistent tracking with its attention.

In this work, we explore the use of attention-based image-to-text
models \cite{xu2015show} for the problem of generating structured
markup. We consider whether a supervised model can learn to produce
correct presentational markup from an image, without requiring a
textual or visual grammar of the underlying markup language. Our
model incorporates a
multi-layer convolutional network over the image with an
attention-based recurrent neural network decoder. To adapt this model
to the OCR problem and capture the document's layout, we also
incorporate a new source encoder layer in the form of a multi-row
recurrent model as part of the encoder. 

Our modeling contributions are twofold. First, we show that
assumptions like the left-to-right ordering inherent in CTC-based
models are not required for neural OCR, since general-purpose encoders
can provide the necessary tracking for accurate attention (example
shown in Figure~\ref{fig:visattn}). Second, in order to reduce
attention computation overhead, we introduce a novel two-layer
hard-soft approach to attention, which we call coarse-to-fine
attention, inspired by coarse-to-fine inference
\cite{raphael2001coarse} from graphical models.\footnote{Note that
  ideas with the same name have been proposed in previous work
  \cite{mei2016}, albeit in a different formulation without the goal
  of reducing computation.} Sparse memory and conditional computation
with neural networks have also been explored with various levels of
success in several previous works \cite{bengio2015conditional,
  Shazeer2017, rae2016sparsememory, andrychowicz2016HAM}. We
demonstrate here that this coarse-to-fine method, when trained with
REINFORCE, significantly reduces the overhead of attention, and leads
to only a small drop in accuracy.

To make these experiments possible, we also construct a new public
dataset, \textsc{Im2Latex-100k}, which consists of a large collection
of rendered real-world mathematical expressions collected from
published articles\footnote{This dataset is based on the challenge
  originally proposed as an OpenAI Request for Research under the
  title Im2Latex.}. This dataset provides a challenging test-bed for
the image-to-markup task based on reconstructing mathematical markup from
rendered images, originally written by scientists. A model is trained to generate LaTeX
markup with the goal of rendering to the exact source image.

Experiments compare the output of the model with several research and
commercial baselines, as well as ablations of these models. The full
system for mathematical expression generation is able to reproduce the same image on more than 75\%
of real-world test examples. Additionally, the use of a multi-row
encoder leads to a significant increase in performance. We also
experiment with training on a simulated handwritten version of the dataset 
to recognize handwritten textual expressions. Even with only a small
in-domain training set, the model is able to produce over 30\% exact match
output. All data, models, and evaluation scripts are publicly
available at \url{http://lstm.seas.harvard.edu/latex/}.

\section{Problem: Image-to-Markup Generation}
\label{sec:background}

We define the image-to-markup problem as converting a rendered source image
to target presentational markup that fully describes both its content and
layout. The source, $\mathbf{x}$, consists of an image. The target, $\mathbf{y}$, consists of a
sequence of tokens $y_1, y_2, \cdots,y_T$ where $T$ is the length
of the output, and each $y$ is a token in the markup language.  The rendering is defined by a possibly unknown, many-to-one,
compile function,
$\operatorname{compile}$. In practice
this function may be quite complicated, e.g a browser, or
ill-specified, e.g. the LaTeX language.

The supervised task is to learn to approximately invert the compile
function using supervised examples of its behavior.  We assume that we
are given instances
$(\mathbf{x}, \mathbf{y})$,
with possibly differing dimensions and that,
$\operatorname{compile}(\mathbf{y}) \approx \mathbf{x}$, for all
training pairs $(\mathbf{x}, \mathbf{y})$ (assuming possible noise).

At test time, the system is given a raw input $\mathbf{x}$ rendered
from ground-truth $\mathbf{y}$. It generates a hypothesis
$\hat{\mathbf{y}}$ that can then be rendered by the black-box function
$\hat{\mathbf{x}} = \operatorname{compile}(\hat{\mathbf{y}})$.  Evaluation
is done between $\hat{\mathbf{x}}$ and $\mathbf{x}$, i.e.  the aim is
to produce similar rendered images while $\hat{\mathbf{y}}$ may or may not be similar to the
ground-truth markup $\mathbf{y}$.

\section{Model}
\begin{figure}[t]
    \centering
        \includegraphics[width=0.77\linewidth]{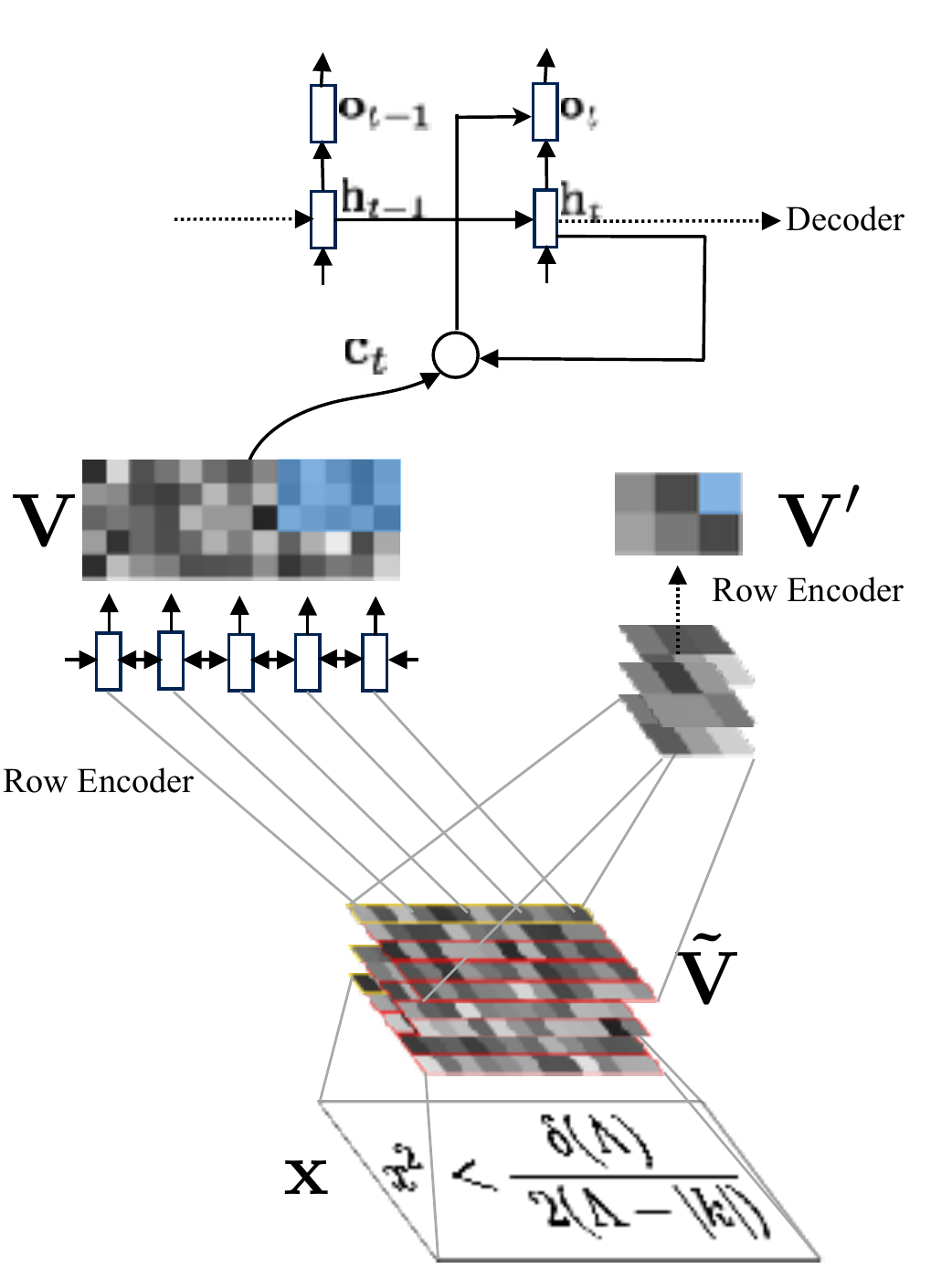}
        \caption{\small Network structure. Given an input image, a CNN
          is applied to extract a feature map
          $\mathbf{\tilde{V}}$. For each row in the feature map, we
          employ an RNN to encode spatial layout
          information. The encoded \textit{fine} features $\mathbf{V}$
          are then used by an RNN decoder with a visual attention
          mechanism to produce final outputs. For clarity we only show
          the RNN encoding at the first row and the decoding at one
          step.  In Section 4, we consider variants of the model where
          another CNN and row encoder are applied to the feature map
          to extract \textit{coarse} features $\mathbf{V}'$, which are
          used to select a support region in the fine-grained
          features, as indicated by the blue masks. }
    \label{fig:network}
    \vspace{-10pt}
\end{figure}

Contrary to most past work on neural OCR, our model uses a full grid
encoder over the input image, so that it can support non left-to-right
order in the generated markup.  The base model is adapted from the
encoder of \citet{xu2015show} developed for image
captioning. Notably, though, our model also includes a row encoder 
which helps the performance of the system. 

The model first extracts image features using a convolutional neural
network (CNN) and arranges the features in a grid. Each row is then
\textit{encoded} using a recurrent neural network (RNN). These encoded
features are then used by an RNN decoder with a visual attention
mechanism. The decoder implements a conditional language model over
the vocabulary, and the whole model is trained to maximize
the likelihood of the observed markup. The full structure is
illustrated in Figure~\ref{fig:network}.

\paragraph{Convolutional Network}
\label{sec:cnn}

The visual features of an image are extracted with a multi-layer
convolutional neural network interleaved with max-pooling layers.
This network architecture is now standard; we model it specifically
after the network used by \citet{shi2015end} for OCR
from images (specification is given in
Table~\ref{tab:cnn}). Unlike some recent OCR
work~\cite{jaderberg2015deep,lee2016recursive}, we do not use final
fully-connected layers \cite{ioffe2015batch}, since we want to
preserve the locality of CNN features in order to use visual
attention. The CNN takes the raw input  and
produces a feature grid $\mathbf{\tilde{V}}$ of size $D \times H \times W$,
where $D$ denotes the number of channels and $H$ and $W$ are the
resulted feature map height and width.

\paragraph{Row Encoder}
\label{sec:encoder}

In image captioning, the CNN features are used as is. For OCR, however, it is
important for the encoder to localize the relative positions
within the source image. In past work this localization has been handled by
CTC, which in effect partitions the source into regions. We instead
implicitly allow the encoder to localize its input by running RNNs
over each of the rows of CNN features. This extension turns out 
to be crucial for performance.

Formally, a recurrent neural network (RNN) is a parameterized function
$\mathbf{RNN}$ that recursively maps an input vector and a hidden
state to a new hidden state. At time $t$, the hidden state is updated
with an input $\mathbf{v}_t$ in the following manner:
$\mathbf{h}_{t} = \mathbf{RNN}(\mathbf{h}_{t-1}, \mathbf{v}_t;
\theta)$,
with $\mathbf{h}_{0}$ an initial state. In practice there are many
different variants of $\mathbf{RNN}$; however, long short-term
memory networks (LSTMs) \cite{hochreiter1997long} have been shown to
be very effective for most NLP tasks. For simplicity we will describe
the model as an RNN, but all experiments use LSTM networks.
 
In this model, the new feature grid $\mathbf{{V}}$ is created
from $\mathbf{\tilde{V}}$ by running an RNN across each row of that input.
Recursively for all rows $h \in \{1, \ldots, H\}$ and columns
$w \in \{1, \ldots, W\}$, the new features are defined as
$\mathbf{{V}}_{hw} = \mathbf{RNN}(\mathbf{{V}}_{h, w-1},
\mathbf{V}_{hw})$.
In order to capture the sequential order information in vertical
direction, we use a trainable initial hidden state $\mathbf{{V}}_{h, 0}$ for each row, which
we refer to as positional embeddings.

\paragraph{Decoder}
\label{sec:attention}
The target markup tokens $\{y_t\}$ are then generated by a decoder
based only on the grid $\mathbf{{V}}$. The
decoder is trained as a conditional language model to give the
probability of the next token given the history and the annotations.
This language model is defined on top of a decoder RNN,
\[p(y_{t+1} | y_1, \ldots, y_{t}, \mathbf{{V}}) =
  \text{softmax}(\mathbf{W}^{out} \mathbf{o}_{t}) \]
where $\mathbf{o}_t = \tanh (\mathbf{W}^c [\mathbf{h}_t;\,
\mathbf{c}_t])$ and $\mathbf{W}^{out}, \mathbf{W}^c$ are learned linear transformations. The vector
$\mathbf{h}_t$ is used to summarize the decoding history:
$\mathbf{h}_{t} = \mathbf{RNN}(\mathbf{h}_{t-1}, [y_{t-1};
\mathbf{o}_{t-1}])$.
The context vector $\mathbf{c}_t$ is used to capture
the context information from the annotation grid. 
We describe how to compute $\mathbf{c}_t$ in the next section.

\section{Attention in Markup Generation}

The accuracy of the model is dependent on being able to track the next
current position of the image for generating markup, which is conveyed
through an attentive context vector $\mathbf{c}_t$.  Formally, we
define a latent categorical variable
$z_t \in \{1,\cdots,H\}\times\{1,\cdots,W\}$ to denote which cell the model 
is attending to.  If we assume access to an attention distribution
$z_t\sim p(z_t)$, then the context is defined as an expectation of
source side features:
\[    \mathbf{c}_t = \sum_{h,w} p(z_{t} = (h,w))\mathbf{{V}}_{hw} \]
In practice, the attention distribution is parameterized as part of the model.
We consider three forms of attention: standard, hierarchical, and coarse-to-fine.

\paragraph{Standard Attention}
In standard attention \cite{bahdanau2014neural}, we use a neural network to approximate the attention distribution $p(z_t)$:
\begin{align*}
 p(z_t) = \text{softmax}(a(\mathbf{h}_{t}, \{\mathbf{{V}}_{hw}\}))
\end{align*}
where $a(\cdot)$ is a neural network to produce unnormalized attention weights.
Note there are different choices for $a$ -- we follow past empirical work and use
$a_{t, h, w} =\beta^T \text{tanh}(W_1\mathbf{h}_t+
W_2\mathbf{{V}}_{hw})$
\cite{luong2015effective}.  

Figure~\ref{fig:visattn} shows an example of the attention
distribution at each step of the model.  
Note several key properties about the attention distribution for the image-to-text
problem. 1) It is important for the grid to be relatively small for
attention to localize around the current symbol. For this reason we use a
\textit{fine} grid with a large $H$ and $W$. 2) In practice, the
support of the distribution is quite small as a single markup symbol
is in a single region. 3) As noted above, attention is run every time
step and requires an expectation over all cells.  
Therefore the decoding complexity of such an attention mechanism is
$O(T H W)$, which can be prohibitive when applied to large images.

\paragraph{Hierarchical Attention}
When producing a target symbol from an image, we can infer the rough
region where it is likely to appear from the last generated symbol
with high probability. In addition to the \textit{fine} grid, we
therefore also impose a grid over the image, such that
each cell belongs to a larger region. When
producing the markup, we first attend to the coarse grid to get the
relevant coarse cell(s), and then attend to the inside fine cells to
get the context vector, a method known as \emph{hierarchical
  attention}.

For this problem, define $\mathbf{V}'$ as a coarse grid of size
${H'} \times {W'}$, which we construct by running additional
convolution and pooling layers and row encoders on top of
$\mathbf{\tilde{V}}$.  We also introduce a latent attention variable
$z_t'$ that indicates the parent level cell of the attended cell, and
write $p(z_t)$ as $p(z_t) = \sum_{z_t'} p(z_t') p(z_t|z_t')$, where we
first generate a coarse-level cell $z_t'$ followed by a fine-level
cell $z_t$ only from within it.

We parameterize $p(z_t')$ and $p(z_t|z_t')$ as part of the model.  For
$p(z_t')$, we employ a standard attention mechanism over
$\mathbf{{V}}'$ to approximate the probability in time $O({H' W'})$.
For the conditional $p(z_t | z_t')$, we also employ a standard
attention mechanism to get as before, except that we only consider the
fine-level cells within coarse-level cell $z_t'$. Note that computing
$p(z_t | z_t')$ takes time $O(\frac{H}{H'} \frac{W}{W'})$. However to
compute the $p(z_{t})$ even with this hierarchical attention, still
requires $O(H W)$ as in standard attention.

\paragraph{Coarse-to-Fine Attention}
Ideally we could consider a reduced set of possible coarse cells in
hierarchical attention to reduce time complexity. Borrowing the name
coarse-to-fine inference \cite{raphael2001coarse} we experiment with
methods to construct a coarse attention $p(z_t')$ with a sparse
support to reduce the number of fine attention cells we consider. We
use two different approaches for training this sparse coarse
distribution.

For the first approach we use sparsemax attention
\cite{martins2016softmax} where instead of using a softmax for
$p(z_t')$ at the coarse-level, we substitute a Euclidean
projection onto the simplex.  The sparsemax function is defined as,
$\text{sparsemax}(\mathbf{p}) = \operatorname{argmin}_{\mathbf{q}\in
  \Delta^{K-1}}\|\mathbf{q} -\mathbf{p}\|_2$,
where $\Delta^{K-1}$ is the probability simplex and $K$ denotes the
number of classes. The sparsemax function can be computed efficiently
and as a projection and can be shown to produce a sparser output than
the standard softmax. If
there are $K^+$ nonzero entries returned by sparsemax, then the
attention time complexity for one step is
$ O(H' W' + K^+\frac{H}{H'}\frac{W}{W'})$.  In practice, we find $K^+$
to be suitably small.

For the second approach we use ``hard'' attention for $z_t'$, an
approach which has been shown to work in several image tasks
\cite{xu2015show, mnih2014visualattention, ba2015visualattention}.
Here we take a hard sample from $p(z_t')$ as opposed to considering the
full distribution.  Due to this stochasticity, the objective is no
longer differentiable. However, stochastic networks can be trained
using the REINFORCE algorithm \cite{williams1992simple}. We pose the
problem in the framework of reinforcement learning by treating $z_t'$
as our agent's stochastic action at time $t$ and the log-likelihood of
the symbol produced as the reward $r_t$. We aim to maximize the total
expected reward $\mathbb{E}_{z_t'}[\sum_{t=1}^T r_t]$, or equivalently
minimize the negative expected reward as our loss.

For parameters $\theta$ that precede the nondifferentiable $z_t'$ in
the stochastic computation graph, we backpropagate a gradient of the
form
$r_t \cdot \frac{\partial \log p(z_t' ; \theta)}{\partial \theta}$.
This gives us an unbiased estimate of the loss function gradient
\cite{schulman2015gradient}. Since our decoder RNN takes previous
context vectors as input at each time step, each action $z_t'$
influences later rewards $r_t, r_{t+1}, \ldots, r_T$. Hence, we assume
a multiplicative discount rate of $\gamma$ for future rewards, and we
use the reward $\tilde{r}_t = \sum_{s=t}^T \gamma^s r_s$ in place of
$r_t$.

In practice, this gradient estimator is noisy and slow to converge. Following \citet{xu2015show}, we include a moving average reward baseline for each timestep $t$ that we update as $b_t \gets \beta b_t + (1-\beta) \tilde{r}_t$, where $\beta$ is a tunable learning rate. We subtract these baselines from our rewards to reduce the variance, giving final gradient update
\[\frac{\partial \mathcal{L}}{\partial \theta} = (\tilde{r}_t - b_t) \cdot \frac{\partial \log p(z_t' ; \theta)}{\partial \theta}.\]
At train time, we sample $z_t'$ and update the network with stochastic
gradients. At test time, we take an argmax over the coarse-level
attentions to choose $z_t'$. The attention time
complexity for a single time step is thus
$O(\frac{H}{H'}\frac{W}{W'} + H'W')$. If we take $H'=\sqrt{H}$,
$W'=\sqrt{W}$, we get $O(\sqrt{HW})$ attention complexity per decoding
step.

\begin{table}
\centering
\scriptsize
\begin{tabular}{ll}
  \toprule
     Conv & Pool \\
  \midrule  
  coarse features \\
     c:512, k:(3,3), s:(1,1), p:(1,1), bn & - \\
     c:512, k:(3,3), s:(1,1), p:(1,1), bn & po:(4,4), s:(4,4), p:(0,0)\\\hline
     fine features \\
     c:512, k:(3,3), s:(1,1), p:(0,0), bn & -\\
     c:512, k:(3,3), s:(1,1), p:(1,1), bn &  po:(2,1), s:(2,1), p:(0,0)\\
     c:256, k:(3,3), s:(1,1), p:(1,1) &    po:(1,2), s:(1,2), p(0,0)\\
     c:256, k:(3,3), s:(1,1), p:(1,1), bn & -\\
     c:128, k:(3,3), s:(1,1), p:(1,1) &  po:(2,2), s:(2,2), p:(0,0)\\

     c:64, k:(3,3), s:(1,1), p:(1,1)  & po:(2,2), s:(2,2), p(0,0)\\

  \bottomrule
\end{tabular}
 
\caption{\label{tab:cnn} \footnotesize CNN specification. `Conv`: convolution layer, `Pool: max-pooling layer. `c': number of filters, `k': kernel size, `s': stride size, `p': padding size, `po': , `bn': with batch normalization. The sizes are in order (height, width).}
\vspace{-8pt}
\end{table}

\section{Dataset Construction}

To experiment on this task we constructed a new public dataset,
\textsc{im2latex-100k}, which collects a large-corpus of real-world
mathematical expressions written in LaTeX. This dataset provides a
difficult test-bed for learning how to reproduce naturally occurring
rendered LaTeX markup. 

\paragraph{Corpus} 
The \textsc{im2latex-100k} dataset provides 103,556 different LaTeX
math equations along with rendered pictures. We extract formulas by
parsing LaTeX sources of papers from tasks I and II of the 2003 KDD cup \cite{kddcup}, which contain over 60,000 papers. 

We extract formulas from the LaTeX sources with regular expressions, and only keep matches whose number of characters fall in the
range from 40 to 1024 to avoid single symbols or text
sentences.  With these settings we extract over 800,000 different formulas, out of which around 100,000 are
rendered in a vanilla LaTeX environment. Rendering is done with
\textit{pdflatex}\footnote{LaTeX (version 3.1415926-2.5-1.40.14)} and
formulas that fail to compile are excluded. The rendered PDF
files are then converted to PNG format\footnote{ We use the \textit{ImageMagick} \texttt{convert} tool with parameters
  \texttt{-density 200 -quality 100}}.  The final dataset we provide
contains 103,556 images of resolution 1654 $\times$ 2339, and the
corresponding LaTeX formulas.

The dataset is separated into training set (83,883 equations),
validation set (9,319 equations) and test set (10,354 equations) for
a standardized experimental setup. 
The LaTeX formulas range from 38 to 997 characters, with mean 118 and median 98.

\paragraph{Tokenization}

Training the model requires settling on a token set. One
option is to use a purely character-based model. While this method
requires fewer assumptions, character-based models would be significantly more memory intensive than
word-based models due to longer target sequences. Therefore original markup is simply split
into minimal meaningful LaTeX tokens, e.g. for observed characters,
symbols such as \verb|\sigma|, modifier characters such as \verb|^|,
functions, accents, environments, brackets and other miscellaneous
commands. 

Finally we note that naturally occurring LaTeX contains many different
expressions that produce identical output. We therefore experiment
with an optional normalization step to eliminate spurious ambiguity
(prior to training). For normalization, we wrote a LaTeX
parser\footnote{Based on KaTeX parser
  \url{https://khan.github.io/KaTeX/}} to convert the markup to an
abstract syntax tree. We then apply a set of safe normalizing tree
transformation to eliminate common spurious ambiguity, such as fixing
the order of sub-super-scripts and transforming matrices to arrays.
Surprisingly we find this additional step gives only a small 
accuracy gain, and is not necessary for strong results.

\begin{figure}[H]
  \centering
  \includegraphics[width=\linewidth]{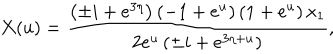}
  \caption{An example synthetic handwritten image from \textsc{im2latex-100k} dataset.}
  \label{fig:synthetic}
\end{figure}

\paragraph{Synthetic Data for Handwriting Recognition}
Our main results focus on rendered markup, but we also considered the
problem of recognizing handwritten math. As there is very little
labeled data for this task, we also synthetized a handwritten corpus
of the \textsc{im2latex-100k} dataset. We created this data set by
replacing all individual symbols with handwritten symbols taken from
Detexify's training
data\footnote{http://detexify.kirelabs.org/classify.html}. We use the
same set of formulas as in the original dataset, but when rendering
each symbol we randomly pick a corresponding handwritten symbol from
Detexify. An example of synthesized handwriting is shown in
Figure~\ref{fig:synthetic}. Note that although the images in this
dataset look like handwritten formulas, they do not capture certain
aspects such as varying baselines \cite{nagabhushan2010tracing}. We
use this dataset as a pretraining step for handwritten formulas
recognition on a small labeled dataset.

\section{Experiments}
\label{sec:exp}

\begin{table*}[t]
  \small
  \centering
  \begin{tabular}{lllcccc}
    \toprule
    \textbf{Dataset} & \textbf{Model} & \textbf{Attention} &  \textbf{BLEU (tok)} & \textbf{BLEU (norm)}  & \textbf{Match} & \textbf{Match (-ws)} \\ 
    \midrule 
    \multirow{9}{*}{Im2latex-100k}
    & \textsc{Infty} & n/a &  51.20 & 66.65 & 15.60 & 26.66 \\
    & \textsc{CTC} & n/a & 39.20 & 30.36 & 7.60 & 9.16 \\
    & \textsc{Caption} & standard & 52.53 & 75.01 & 53.53 & 55.72 \\
    & \textsc{Im2Tex-tok} & standard & 73.71 & 73.97 & 74.46 & 77.04 \\
    & \textsc{Im2Tex} & standard & 58.41 & 87.73 & 77.46 & 79.88\\
     \cmidrule{2-7} 
    & \textsc{Im2Tex} & coarse-only & 53.40 & 79.53 & 44.40 & 48.53\\
    & \textsc{} & hierarchical& 60.32 & 86.21 & 77.39 & 79.63\\
    & \textsc{Im2Tex-C2F} &  hard & 59.96 & 87.07 & 74.90 & 77.07\\ 
    & \textsc{} & sparsemax &  59.38 & 86.97 & 76.15 & 78.10\\
    \midrule \multirow{8}{*}{CROHME14}
    & \textsc{MyScript}* & n/a & - & - & 62.68 & -\\
    & \textsc{UPV} & n/a &  - & - & 37.22 & -\\
    & \textsc{U Nates} & n/a & - & - & 26.06 & -\\
    & \textsc{TUAT} & n/a & - & - & 25.66 & -\\
    & \textsc{Im2Tex} &standard& 50.28 & 68.57 & 38.74 & 39.96\\
     \cmidrule{2-7} 
    & \textsc{Im2Tex}&hierarchical & 47.52 & 64.49 & 35.90 & 36.41\\ 
    & \textsc{Im2Tex-C2F} &hard& 48.02 & 64.42 & 33.87 & 35.09\\ 
    & \textsc{} &sparsemax&  47.65 & 64.26 & 34.28 & 35.40\\
    \midrule \multirow{8}{*}{CROHME13}
    & \textsc{MyScript}* & n/a &  - & - & 60.36 & -\\
    & \textsc{U Valencia} & n/a &  - & - & 23.40 & -\\
    & \textsc{TUAT} & n/a & - &  - & 19.97 & -\\
    & \textsc{USP} & n/a & - &  - & 9.39 & -\\
    & \textsc{Im2Tex} &standard& 44.51 & 60.84 & 33.53 & 34.72\\
     \cmidrule{2-7} 
    & \textsc{Im2Tex} &hierarchical&  43.65 & 59.70 & 29.81 & 30.85\\ 
    & \textsc{Im2Tex-C2F} &hard & 43.09 & 57.18 & 29.51 & 31.00\\ %
    & \textsc{} &sparsemax& 44.18 & 58.13 & 28.32 & 29.51\\
      \bottomrule
  \end{tabular}  \caption{\label{tab:main} \small [Top] Main experimental results on the \textsc{im2latex-100k} dataset. Reports the BLEU score compared to the tokenized formulas (BLEU (tok)), BLEU score compared to the normalized formulas (BLEU (norm)), exact match accuracy, and exact match accuracy after deleting whitespace columns. All systems except \textsc{Im2Tex-Tok} are trained on normalized data. [Bottom] Results on the CROHME handwriting datasets. We list the best 4 systems in 2013 and 2014 competition: MyScript, U Valencia, TUAT, USP, and MyScript, UPV, U Nates, TUAT. All Im2Tex systems use out-of-domain synthetic data as well as the small given training set. *Note that the proprietary MyScript system uses a large corpus of private in-domain handwritten training data.}
\vspace*{-0.00cm}
\end{table*}

Experiments compare the proposed model, which we refer to as
\textsc{Im2Tex} to classical OCR baselines, neural models, and
model ablations on the image-to-LaTeX task.  We also compare the
proposed model against commercial, OCR-based mathematical expression
recognition system InftyReader. InftyReader is an implementation of
the \textsc{Infty} system of \cite{suzuki2003infty}, 
combining symbol recognition and structural analysis phases.

For neural models, a natural comparison is to standard image
captioning approaches \cite{xu2015show}, and CTC-based approaches
\cite{shi2016end}.  We simulate the image captioning setup with a
model \textsc{Caption} which removes the row encoder, i.e. replacing
$\mathbf{{V}}$ with $\mathbf{\tilde{V}}$, and increases the number of
CNN filters such that the number of parameters is the same. For CTC
we use the implementation of \citet{shi2016end}, designed for natural 
image OCR.

To better understand the role of attention in the model, we run
several baseline experiments with different attention styles.  To
examine if fine-level features are necessary, we experiment with a
standard attention system with the coarse feature maps only
(coarse-only) and also with a two-layer hierarchical model.
Additionally we experiment with different coarse-to-fine (C2F)
mechanisms: hard reinforcement learning, and
sparsemax.

Finally, we run additional experiments comparing our approach to other
models for handwritten mathematical expressions on the CROHME 2013 and
2014 shared tasks. The training set is same for both years, consisting of 8,836 training expressions (although teams also used
external data). The dataset is in a different domain from our rendered
images and is designed for stroke-based OCR. To handle these
differences, we employ two extensions: (1) We convert the data to
images by rendering the strokes and also augment data by
randomly resizing and rotating symbols, (2) We also employ the
simulated \textsc{im2latex-100k} handwriting dataset to pretrain a
large out-of-domain model and then fine-tune it on this CROHME dataset.

Our core evaluation method is to check the accuracy of the rendered
markup output image $\hat{\mathbf{x}}$ compared to the true image
$\mathbf{x}$. The main evaluation reports exact match rendering
between the gold and predicted images, and we additionally check the
exact match accuracy with the original image as well as the value
after eliminating whitespace columns.\footnote{ In practice we found
  that the LaTeX renderer often misaligns identical expressions by
  several pixels. To correct for this, only misalignments of $\geq$ 5
  pixels wide are ``exact'' match errors. } We also include standard
intrinsic text generation metrics, conditional language model
perplexity and BLEU score \cite{papineni2002bleu}, on both tokenized
and normalized gold data.

\paragraph{Implementation Details}
 
The CNN specifications are summarized in Table~\ref{tab:cnn}. Note that $\frac{H}{H'}=\frac{W}{W'}=4$. The
model uses single-layer LSTMs for all RNNs. We use a bi-directional
RNN for the encoder. The hidden state of the encoder RNN is of size
256, decoder RNN of 512, and token embeddings of size 80. The model
with standard attention has 9.48 million parameters, and the models
with hierarchical or coarse-to-fine attention have 15.85 million
parameters due to the additional convolution layers and row encoders.  We
use mini-batch stochastic gradient descent to learn the parameters.

For the standard attention models, we use batch size of 20. The
initial learning rate is set to 0.1, and we halve it once the
validation perplexity does not decrease. We train the model for 12
epochs and use the validation perplexity to choose the best
model. For the
hierarchical and coarse-to-fine attention models, we use batch size of
6. For hard attention, we use the pretrained weights of
hierarchical to initialize the parameters. Then we use
initial learning rate 0.005, average reward baseline learning rate
$\beta = 0.01$, reward discount rate $\gamma = 0.5$.

The complete model is trained end-to-end to maximize the likelihood of
the training data. Beyond the training data, the model is
given no other information about the markup language or the generating
process. To generate markup from unseen images, we use beam search with beam size 5 at
test time. No further hard constraints are employed.

The system is built using Torch \cite{collobert2011torch7} based on
the OpenNMT system \cite{klein2017opennmt}. Experiments are run on a 12GB Nvidia Titan
X GPU (Maxwell).

Original images are cropped to only the formula area, and padded with 8 pixels to the top, left, right and bottom.
For efficiency we downsample all images to half of their original sizes.
To facilitate batching, we group images into similar sizes and pad with whitespace.\footnote{ Width-Height groups used are (128, 32), (128, 64), (160, 32), (160, 64), (192, 32), (192, 64), (224, 32), (224, 64), (256, 32), (256, 64), (320, 32), (320, 64), (384, 32), (384, 64), (384, 96), (480, 32), (480, 64), (480, 128), (480, 160).}
All images of larger sizes, LaTeX formulas with more than 150 tokens, or those that cannot be parsed are ignored during training and validation, but included during testing. 

\section{Results}

The main experimental results, shown at the top of Table~\ref{tab:main}, compare
different systems on the image-to-markup task. The
\textsc{Infty} system is able to do quite well in terms of text
accuracy, but performs poorly on exact match image
metrics. The poor results of the neural \textsc{CTC} system validate
our expectation that the strict left-to-right order assumption is
unsuitable in this case.  Our reimplementation of image captioning
\textsc{Caption} does better, pushing the number above
50\%. Our standard attention system \textsc{Im2Tex} with RNN encoder
increases this value above 75\%, achieving high accuracy on this
task. The LaTeX normalizer provides a few points of accuracy gain and
achieves high normalized BLEU. This indicates that the decoder LM is
able to learn well despite the ambiguities in real-world LaTeX.
\begin{table}[!tp]
    \centering
        \small
    \begin{tabular}{llcccc}
      \toprule
      
      Model & Ablation & Train &  Test &Match\\
      \midrule
      \textsc{NGram} & & 5.50 & 8.95& - \\      
      \textsc{LSTM-LM} & -Enc & 4.13 &   5.22& -\\
      \textsc{Im2Tex}& -RowEnc & 1.08 &  1.18&53.53\\
      \textsc{Im2Tex}& -PosEmbed & 1.03 &  1.12&76.86\\
      \textsc{Im2Tex} & & 1.05 & 1.11&77.46 \\
      \textsc{Im2Tex-C2F} (hard) & & 1.05 & 1.15&74.90 \\
      \bottomrule
    \end{tabular}
    \caption{\label{tab:result} \small Image-to-LaTeX ablation experiments. Compares simple LM approaches and versions of the full model on train and test perplexity, and
 image match accuracy.}
\vspace*{-0.31cm}
\end{table}

We next compare the different hierarchical and coarse-to-fine
extensions to the system. We first note that the use of the
coarse-only system leads to a large drop in accuracy, indicating that
fine attention is crucial to performance.  On the other hand, the high
performance of hierarchical indicates that two layers of
soft-attention do not hurt the performance of the model.
Table~\ref{tab:num_lookup} shows the average number of cells being
attended to at both the coarse and fine layers by each of the
models. Both the hard REINFORCE system and sparsemax reduce lookups at
a small cost in accuracy. Hard is the most aggressive, selecting a
single coarse cell. Sparsemax achieves higher accuracy, at the cost of
selecting multiple coarse cells. Depending on the application, these
are both reasonable alternatives to reduce the number of lookups in
standard attention.

Our final experiments look at the CROHME 2013 and 2014 datasets, which
were designed as a stroke recognition task, but are the closest
existing dataset to our task. For this dataset we first train with our synthetic handwriting dataset and then fine-tune on the CROHME
training set. We find our models achieve comparable performance to
all best systems excepting \text{MyScript}, a commercial system with
access to additional in-domain data. Note that our synthetic dataset
does not contain variation in baselines, font sizes, or other noise,
which are common in real data. We expect increased performance from the
system when trained with well-engineered data. For these datasets we
also use the hierarchical and coarse-to-fine models, and find that
they are similarly effective.  Interestingly, contrary to the full
data for some problems hard performs better than sparsemax.

\paragraph{Analysis}

\begin{figure}
  \centering
  \includegraphics[width=\linewidth]{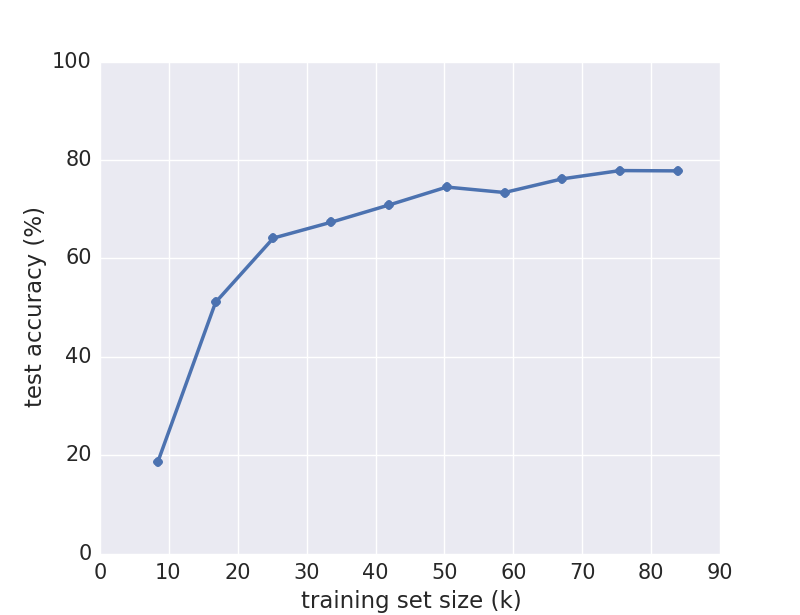}
  
  \caption{\small Test accuracy (Match) of the model w.r.t. training set size.}
  \label{fig:train_size}
  \vspace*{-0.5cm}
\end{figure}

To better understand the contribution of each part of the standard
\textsc{Im2Tex} model, we run ablation experiments removing different
features from the model, which are shown in
Table~\ref{tab:result}. The simplest model is a basic
(non-conditional) \textsc{NGram} LM on LaTeX which achieves a
perplexity of around 8. Simply switching to an LSTM-LM reduces the
value to 5, likely due to its ability to count parentheses and
nesting-levels. These values are quite low, indicating strong
regularity just in the LaTeX alone. Adding back the image data with a
CNN further reduces the perplexity down to 1.18. Adding the encoder
LSTM adds a small gain to 1.12, but makes a large difference in final
accuracy. Adding the positional embeddings (trainable initial states
for each row) provides a tiny gain. Hard attention leads to a
small increase in perplexity.  We also consider the effect of training
data on performance. Figure~\ref{fig:train_size} shows accuracy of the
system with different training set size using standard attention.  As
with many neural systems, the model is quite data hungry. In order for
the model to reach $\ge 50\%$ accuracy, at least 16k training examples
are needed.

\begin{figure}[H]
  \centering
  \includegraphics[width=\linewidth]{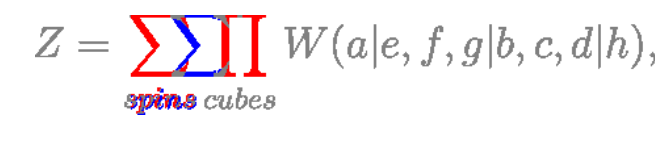}
  \includegraphics[width=\linewidth]{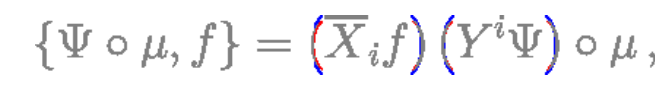}
  \includegraphics[width=\linewidth]{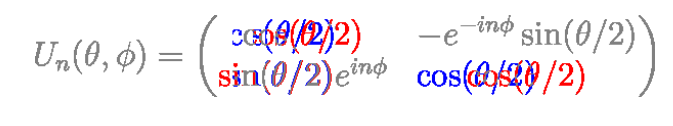}
  \includegraphics[width=\linewidth]{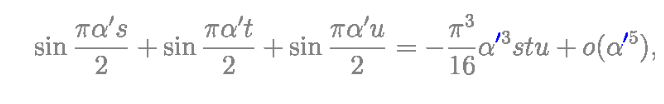}
  \caption{\small Typical reconstruction errors on aligned images. Red denotes gold image and blue denotes generated image.}
  \label{fig:errors}
  \vspace*{-0.3cm}
\end{figure}

Finally Figure~\ref{fig:errors} illustrates several common
errors. Qualitatively the system is quite accurate on difficult LaTeX
constructs. Typically the structure of the expression is preserved
with one or two symbol recognition errors. We find that the most
common presentation-affecting errors come from font or sizing issues,
such as using small parentheses instead of large ones, using standard
math font instead of escaping or using \textit{mathcal}.

\begin{table}[!t]
    \small
    \centering
    \begin{tabular}{llccc}
      \toprule
      Model & Attn & \# C & \# F  &  Match\\
      \midrule
      \textsc{Im2Tex}    & standard     & 0 & 355 & 77.46\\
       & coarse-only       & 22   & 0 & 44.40\\
      & hierarchical     & 22 & 355 & 77.39\\
      \textsc{Im2Tex-C2F} & hard     & 22  & 16 &   74.90\\
      & sparsemax &  22  & 74 &  76.15\\
      \bottomrule
    \end{tabular}
    \caption{ \label{tab:num_lookup} \small Average number of coarse (\#C) and fine (\#F) attention computations for all models throughout the test set. 
      standard and hierarchical provide an upper-bound and coarse-only a lower-board, whereas hard always 
      does the minimal $4\times 4 = 16$ fine lookups. Test accuracy is shown for ease of comparison.}
      \vspace{-9pt}
\end{table}

\section{Conclusion}
We have presented a visual attention-based model for OCR of
presentational markup. We also introduce a new dataset
\textsc{im2latex-100k} that provides a test-bed for this task. In
order to reduce the attention complexity, we propose a coarse-to-fine
attention layer, which selects a region by using a coarse view of the
image, and use the fine-grained cells within.  These contributions
provide a new view on the task of structured text OCR, and show
data-driven models can be effective without any knowledge of the
language. The coarse-to-fine attention mechanism is general and
directly applicable to other domains, including applying the proposed
coarse-to-fine attention layer to other tasks such as document
summarization, or combining the proposed model with neural inference
machines such as memory networks.

\begin{small}

\subsection*{Acknowledgements}
We would like to thank Daniel Kirsch for providing us Detexify data,
and Sam Wiseman and Yoon Kim for the helpful feedback on this paper.
This research is supported by a Bloomberg Data Science Research Award.
\end{small}

\newpage

\bibliography{main}
\bibliographystyle{icml2017}

\end{document}